\documentclass{article}

\usepackage{PRIMEarxiv}

\usepackage[utf8]{inputenc} 
\usepackage[T1]{fontenc}    
\usepackage{hyperref}       
\usepackage{url}            
\usepackage{booktabs}       
\usepackage{amsfonts}       
\usepackage{nicefrac}       
\usepackage{microtype}      
\usepackage{lipsum}
\usepackage{fancyhdr}       
\usepackage{graphicx}       
\graphicspath{{media/}}     

\usepackage{amssymb}
\usepackage{extarrows}
\usepackage{dsfont}

\usepackage[table]{xcolor}
\usepackage{multirow}

\usepackage{amsmath}

\newcommand{\comment}[1]{}

\newcommand{\entity}[1]{\mbox{\textsf{#1}}}

\title{Knowledge Graph Question Answering for Materials Science (KGQA4MAT): Developing Natural Language Interface for Metal-Organic Frameworks 
Knowledge Graph (MOF-KG) Using LLM
}

\author{
  Yuan An, Jane Greenberg, Alex Kalinowski, Xintong Zhao, Xiaohua Hu \\
  College of Computing and Informatics \\
  Drexel University \\
  Philadelphia, PA, 19104\\
  \texttt{\{ya45,jg3243,xz485,ajk437,xh29\}@drexel.edu} \\
   \And
  Fernando J. Uribe-Romo, Kyle Langlois, Jacob Furst \\
  Department of Chemistry \\ 
  University of Central Florida \\ 
  Orlando, FL, USA \\ \texttt{fernando@ucf.edu, \{kylerlanglois,jfurst\}@knights.ucf.edu} \\
  \AND
  Diego A. Gómez-Gualdrón  \\
  Chemical and Biological Engineering \\ 
  Colorado School of Mines \\ 
  Golden, CO, USA \\
  \texttt{dgomezgualdron@mines.edu} \\
}

\begin{document}
\maketitle

\begin{abstract}
We present a comprehensive benchmark dataset for Knowledge Graph Question 
Answering in Materials Science (KGQA4MAT), with a focus on metal-organic 
frameworks (MOFs). A knowledge graph for metal-organic frameworks (MOF-KG) 
has been constructed by integrating structured databases and knowledge 
extracted from the literature. To enhance MOF-KG accessibility for domain experts, 
we aim to develop a natural language interface for querying the knowledge graph. 
We have developed a benchmark comprised of 161 complex questions involving comparison, 
aggregation, and complicated graph structures. Each question is rephrased 
in three additional variations, resulting in 644 questions and 161 KG queries.
To evaluate the benchmark, we have developed a systematic approach for utilizing 
the LLM, ChatGPT,  to translate natural language questions into formal KG queries. We also 
apply the approach to the well-known QALD-9 dataset, demonstrating ChatGPT's 
potential in addressing KGQA issues for different platforms and query languages.
The benchmark and the proposed approach aim to stimulate further research and 
development of user-friendly and efficient interfaces for querying domain-specific 
materials science knowledge graphs, thereby accelerating the discovery of 
novel materials.
\end{abstract}


%
%
\section{Introduction}
\label{sec:introduction}

Metal-organic frameworks (MOFs) are a class of porous materials with high surface area, high porosity, 
and tunable pore size \cite{reticular-chemistry-all-dimensions}, 
which make them attractive for a wide range of applications, including 
gas storage, separation, catalysis, and drug delivery.
Despite their potential, Metal-Organic Frameworks (MOFs) have yet to be fully utilized due 
to a lack of comprehensive and systematic knowledge about their composition, structure, synthesis, 
and properties. 

\begin{figure}[!ht]
	\centering
	\includegraphics[width=0.7\textwidth]{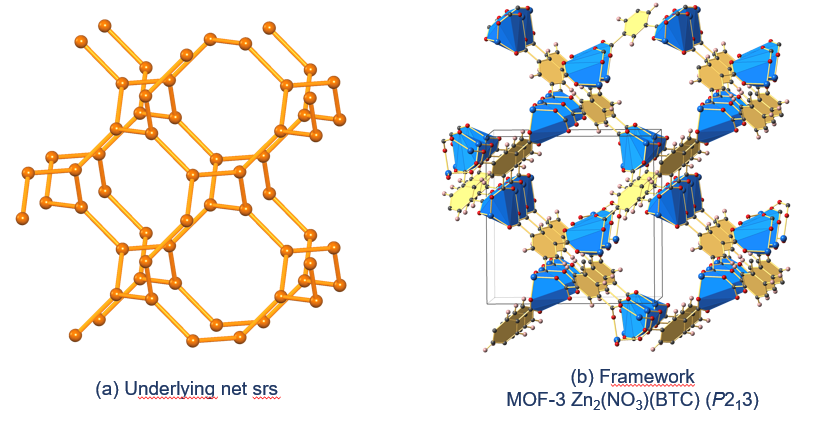}
	\caption{The Underlying Net and Framework of MOF-3}
	\label{fig:underlying-net-MOF-3}
\end{figure}

A MOF consists of metal ions or clusters connected by organic ligands, 
forming a three-dimensional network.
Figure \ref{fig:underlying-net-MOF-3} illustrates 
the porous structure and underlying net of the `MOF-3' that has been extensively  
studied for various applications in the literature \cite{MOF-3}.
The ability to tune MOF properties through composition variations results in a very large number 
of potential MOF candidates. 
There has been a significant push to use computational techniques to identify 
the best MOF structures for various applications. As a result, 
several extensive databases of both 
synthesized and theoretical MOF 
structures \cite{development-CSD-MOF,increasing-topoloical-diversity,understanding-diversity-MOF} 
have been created. In the Cambridge Structural Database (CSD) \cite{development-CSD-MOF}, a leading
database that compiles and maintains crystal structures of small-molecule organic and metal-organic 
compounds, there are more than 100,000 MOF structures that have been documented as of 2022. Additionally, 
more than 500,000 MOFs have been predicted and stored in other databases
\cite{understanding-diversity-MOF,increasing-topoloical-diversity}.
Because it is infeasible to synthesize and test all MOF candidates, 
identifying optimal MOFs for a target application is a extremely challenging task, 
akin to trying to find a needle in a haystack.

The MOF identification problem is further compounded by the fact that 
much of the vital information on MOF synthesis 
is not found in MOF databases, but scattered throughout scientific literature. 
Manually searching through thousands of articles to analyze 
synthesis conditions for a specific MOF structure is both time-consuming and ineffective.
To address the problem, several lines of research have applied 
Natural language processing (NLP) to extract synthesis information from the literature 
\cite{DigiMOF,MOF-synthesis-automatic-dataMining,mining-insights-MOF}.   
However, these studies do not create a comprehensive knowledge graph encompassing all aspects of MOFs. 

To fully utilize the scattered information generated by MOFs research and experiments, it is essential to 
gather and integrate the structural information from MOF databases as well as the synthesis procedures in the 
literature in a easily accessible and searchable place. To this end, we 
integrated the MOF structural information in several MOF databases with the 
synthesis procedures extracted from the literature by a recent 
study \cite{mining-insights-MOF}, resulting in
a comprehensive knowledge graph for metal-organic frameworks called 
MOF-KG \cite{kg-empowered-materials,mof-kg,mof-lama}.

A knowledge graph can be stored in an RDF triple store or in the Neo4j platform. 
Unfortunately, for either type of system, there is a significant technical barrier for 
domain experts to directly utilize the knowledge graph. In particular, it is challenging 
for domain experts to write SPARQL or Cypher queries. To enhance the usability
and accessibility of the MOF-KG for domain experts, we aim to develop an intuitive interface that
can translate domain experts questions in natural language to corresponding SPARQL or Cypher queries.

In this paper, we introduce a benchmark dataset for Knowledge
Graph Question Answering for Materials Science (KGQA4MAT). The benchmark contains 161 complex
natural language questions including comparison, aggregation, and complicated graph structures. 
For each question, we generate three different variations, resulting in total 644 questions. 
Each question is associated with a corresponding formal query on the underlying graph system. 
For the current version of the KGQA4MAT benchmark, we focus on Cypher queries on the Neo4j platform.

To evaluate the benchmark, we have developed a systematic approach of 
utilizing ChatGPT for translating natural language questions into formal KG queries. 
We also apply the approach to the well-known QALD-9 dataset \cite{qald-9} demonstrating ChatGPT's
potential addressing KGQA issues for different platforms and query languages. 
The benchmark, MOF-KG, and the proposed approach aim to stimulate further research and 
development of user-friendly and efficient interfaces for querying domain-specific 
materials science knowledge graphs, thereby accelerating the discovery of novel materials.

The rest of the paper presents the MOF-KG, the KGQA4MAT benchmark, 
the proposed approach, and evaluation results 
for question answering over the MOF-KG.
Section \ref{sec:related_work} presents related work. 
Section \ref{sec:mof_knowledge_graph} describes the MOF-KG.
Section \ref{sec:ask_mof_kg} discusses the challenges of translating NL questions against the MOF-KG.
Section \ref{sec:benchmark} presents the KGQA4MAT benchmark.
Section \ref{sec:approach} describes the proposed approach for KGQA.
Section \ref{sec:evaluation} describes the evaluation results.
Section \ref{sec:discussion} discusses the findings.
Finally, Section \ref{sec:conclusion} concludes the paper and 
presents future directions.

%
%
\section{Related Work}
\label{sec:related_work}

Researchers have begun to develop knowledge graphs in different 
materials areas \cite{MMKG,propnet,Nanomine,MatKG}.
MMKG \cite{MMKG} is a knowledge graph derived from DBpedia and Wikepedia for 
Metallic Materials. Propnet \cite{propnet} is a Python framework to create a knowledge graph by 
connecting materials in existing datasets through their properties. 
NanoMine \cite{Nanomine} is a curated knowledge graph (KG) that has been developed 
for the field of nanocomposite materials science. 
MatKG \cite{MatKG} is a large KG that contains 
general materials science concept and relationships extracted from 4 million 
materials science-related papers. Despite the potential benefits of 
KGs for materials research, there is a lack of KGs specifically tailored to the field of 
metal-organic frameworks (MOFs).

KGs must be easily searchable by domain experts using simple and intuitive methods. 
Most existing materials science KGs were inadequate in this aspect as they require the 
end users to use a highly specialized query languages. 
Knowledge Graph Question Answering (KGQA) aims to enable end users to 
pose natural language questions and receive answers from 
knowledge graphs (KGs) \cite{survey-QA-SW}. 
Benchmark datasets are crucial for evaluating the performance of KGQA systems and fostering 
advancements in the field. The creation of such datasets usually involves the development 
of a set of complex questions with varying degrees of difficulty, 
covering diverse topics and targeting different KG representations. 
Some well-known benchmark datasets include 
QALD series \cite{qald-9}, LC-QuAD series \cite{LC-QuAD,LC-QuAD-2}, 
and SciQA \cite{SciQA}. 
Steinmetz et al. in \cite{what-is-KGQA-benchmark} surveyed
the challenges associated with the creation of many existing KGQA benchmarks.

While the benchmarks such as QALD, LC-QuAD, and SciQA have become influential for 
assessing and advancing the state-of-the art in KGQA, they mainly target at
general knowledge based on large open general knowledge graphs such as DBpedia, 
Wikipedia, and ORKG. There lacks benchmarks on specific domain knowledge graphs such as 
in materials science. As KGQA systems continue to evolve, the development of 
more comprehensive and diverse benchmark datasets will remain essential for future progress.

Several KGQA systems have been developed over the years, each with distinct techniques 
for translating natural language questions into formal SPARQL or Cypher queries. Early 
systems employed rule-based or template-based approaches \cite{comparative-survey-NL-databases}.
Recent KGQA systems have increasingly adopted machine learning and deep learning techniques.
Techniques such as sequence-to-sequence models, 
attention mechanisms, neural machine
translation, and graph neural networks 
\cite{zou14natural,deep-neural-approach-KGQA,learning-to-rank-QG,querying-KG-NL,SGPT-generative} have been employed.
Despite the significant progress in the field, several challenges remain to be 
addressed, such as handling complex queries with multiple relations, 
aggregations, or comparisons. Since the emergence of pre-trained Large Language Models (LLM) and ChatGPT, 
a flurry of
studies 
have been proposed on extracting knowledge graphs \cite{mof-lama,Fang-prompt-design}
and evaluating ChatGPT for KGQA 
\cite{can-chatgpt-understand,how-robust-gpt-3.5,chatgpt-versus-traditional,evaluation-chatGPT-QA}. The KGQA Github 
repository\footnote{https://github.com/KGQA/leaderboard} contains rich KGQA resources
including datasets, systems, and leaderboard. The most recent leaderboard
indicates that KGQA remains a challenging problem demanding 
continued research to develop more accurate and robust systems.

%
%
\section{The MOF Knowledge Graph (MOF-KG)}
\label{sec:mof_knowledge_graph}

The MOF Knowledge Graph (MOF-KG) is a comprehensive representation of knowledge related to 
metal-organic frameworks, encompassing various aspects such as structures, properties, 
and synthesis procedures. To build 
the knowledge graph, we began with defining an ontology and then populated the 
knowledge graph with the data in MOF databases such as 
Cambridge Structural Database (CSD) \cite{development-CSD-MOF} and the extracted 
synthesis procedures from the literature. 

Currently, a universally accepted system for describing metal-organic frameworks (MOFs) 
and related phenomena remains nonexistent. Although a few initiatives have been established 
to standardize the nomenclature \cite{RCSR,terminology-of-MOF}, their scope and applicability 
in integrating MOF information into a comprehensive knowledge graph are limited. Several 
domain-specific materials science ontologies exist, such as Ashino's Materials 
Ontology \cite{Ashino-materials-ontology}, ChEBI (Chemical Entities of Biological 
Interest) \cite{ChEBI}, Elementary Multiperspective Material Ontology (EMMO) \cite{EMMO}, 
Materials Design Ontology (MDO) \cite{MDO}, and the NIST controlled vocabulary \cite{NIST-MaterialsOntology}. 
Additionally, the MatPortal project \cite{MatPortal} serves as an open repository, a
massing over 20 ontologies related to materials science. However, none of these ontologies 
specifically address the distinct concepts and relationships inherent to MOFs.

\begin{figure}[!ht]
	\centering
	\includegraphics[width=1\textwidth]{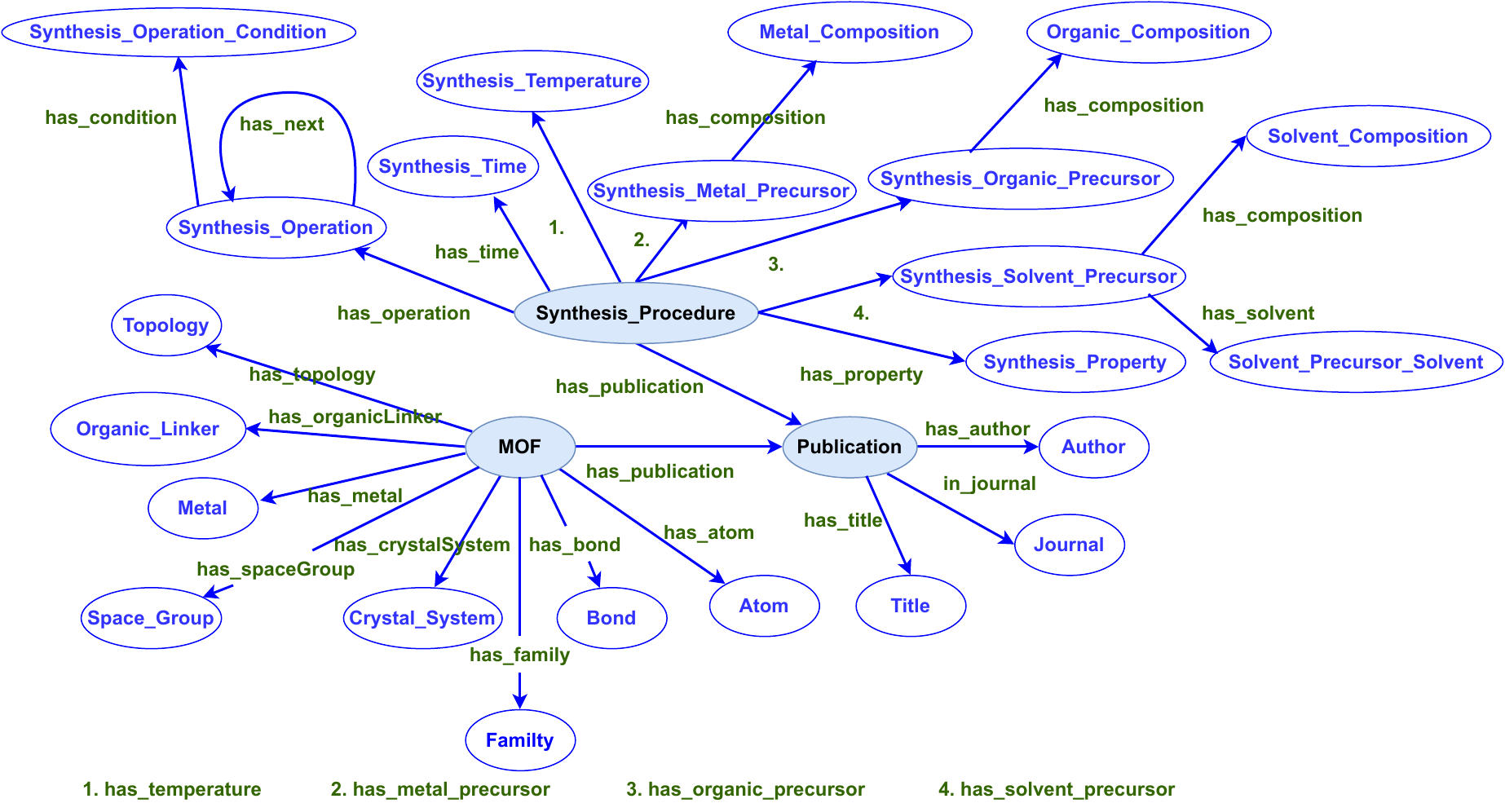}
	\caption{The concepts and relationships in the ontology for the MOF knowledge 
            graph (MOF-KG)}
	\label{fig:MOF-KG-in-Neo4j}
\end{figure}

In our previous work \cite{mof-kg}, an ontology for the Metal-Organic Framework Knowledge 
Graph (MOF-KG) was established, consisting of four primary domains: \emph{synthesis}, 
\emph{structure}, \emph{atomic composition}, and \emph{publication}. This ontology was 
refined through the integration of knowledge from structured MOF databases and the 
extraction of information from relevant literature. Figure \ref{fig:MOF-KG-in-Neo4j} 
illustrates the concepts and relationships within the ontology.

Three key concepts form the foundation of the ontology: \entity{MOF}, \entity{Publication}, 
and \entity{Synthesis\_Procedure}. The \entity{MOF} concept connects to the concepts 
representing MOF structure and atomic composition, while the \entity{Synthesis\_Procedure} 
concept relates to the concepts describing synthesis operations, precursors, and various 
properties. The \entity{Publication} concept links to the concepts regarding authors, 
journals, and titles. Notably, the \entity{Publication} serves as a cut vertex in the 
ontology graph, connecting MOF entities to their corresponding synthesis procedures.

The ontology reflects the absence of naming conventions for MOFs and the organization of the information in the 
original sources. In particular, a database such as the Cambridge Structural Database (CSD) 
assigns a unique reference code, \entity{RefCode}, to a MOF, which is associated with a 
scholarly publication that describes synthesis procedures in the main text or supplemental 
materials. However, when a publication initially presents the synthesis procedures for one or multiple MOFs, 
there are no assigned RefCodes yet. Moreover, the naming of MOFs lacks a 
commonly accepted format, making MOF names unsuitable for identification. Therefore, 
the MOF knowledge graph (MOF-KG) utilizes the publication's globally unique \entity{DOI} 
to bridge the MOFs and the synthesis procedures, thus creating a cut vertex.

The MOF knowledge graph (MOF-KG) is populated by the information extracted from the 
MOF collection\footnote{https://www.ccdc.cam.ac.uk/free-products/csd-mof-collection/} 
in the CSD database, the CoRE MOF dataset\footnote{https://zenodo.org/record/3370144\#.ZDvtN3bMIdV}
created by Chung et al. in \cite{CoREMOF},  the MOFs in the 
Reticular Chemistry Naming and Numbering (RCNN) Database\footnote{https://globalscience.berkeley.edu/database}, 
and the synthesis procedures\footnote{https://doi.org/10.6084/m9.figshare.16902652.v3} 
extracted by Park et al. in the study \cite{mining-insights-MOF}.
The following steps describe how to create the MOF-KG from the sources with an appropriate CSD license:
\begin{enumerate}
    \item  Download the MOF collection, CoRE dataset, RCNN dataset, and the extracted
    MOF synthesis procedures.
    \item Query the CSD crystal database using licensed CSD Python API 
    to extract the information about {crystal system} and 
	{space group}
    \item Apply the CSD ConQuest tool to identify a MOF's {family} by applying the search 
    criteria developed by Moghadam et al. in \cite{targeted-classification-MOFs}. There are six prototypical
	MOF families identified: Zr-oxide nodes (e.g. UiO-66), Cu–Cu paddlewheels (e.g. HKUST-1), ZIF-like, Znoxide
	nodes, IRMOF-like, and MOF-74/CPO-27-like materials.
    \item Leverage the MOFid system\footnote{https://github.com/snurr-group/mofid} developed by 
	Bucior et al. in \cite{identification-schemes-MOFs} to identify a MOF's {metals}, 
	{organic linkers}, and {topology}.
    \item Populate the MOF-KG by directly mapping the extracted data to the ontology in Figure 
    \ref{fig:MOF-KG-in-Neo4j}.
    
\end{enumerate}

The resulting knowledge graph consists of  more than 1.5 million nodes and 
more than 3.7 million relationships between the nodes. The knowledge graph 
is stored in the Neo4j platform, partially due to its intuitive graph representation 
and visualization capabilities.

%
%
\section{Knowledge Graph Query Answering (KGQA) over the MOF-KG}
\label{sec:ask_mof_kg}

Despite the user-friendly graphical user interface provided by the 
Neo4j platform, domain experts still face challenges in querying 
the knowledge graph. One challenge is the need to learn a formal 
query language, such as Cypher. Another challenge involves understanding 
the underlying graph structure and schema. Additionally, formulating 
complex queries to address domain-specific questions can be difficult.
The problem of Knowledge Graph Question Answering (KGQA) over the MOF-KG 
involves enabling users to ask natural language questions about MOFs.
There are multiple obstacles associated with KGQA over the MOF-KG:
\begin{itemize}
    \item \emph{Diverse Query Types}: MOF-KG contains a wide range of information, such as 
    synthesis procedures, structures, atomic compositions, and related publications. 
    Therefore, a KGQA systems should be able to handle a variety of query types, 
    including single- and multi-relation queries, as well as more complex queries 
    involving aggregation, comparison, and negation.

    \item \emph{Ambiguity and Paraphrasing}: Natural language questions can be ambiguous, contain 
    synonyms, or be phrased in multiple ways. A KGQA system should be able to understand 
    and disambiguate user questions and recognize different ways of asking the same question.

    \item \emph{Domain-Specific Language}: MOFs and their associated information are often 
    described using domain-specific jargon, chemical formulas, or notation. A KGQA system 
    should be capable of understanding and handling such domain-specific language to provide accurate answers.
\end{itemize}

Assessing the performance of a KGQA system over the MOF-KG 
can be challenging due to the specialized nature of the domain and 
the lack of comprehensive benchmark datasets. 
Developing suitable evaluation metrics and datasets is essential for 
driving advancements in KGQA systems for MOFs, and materials science more broadly.
In the subsequent sections, we outline a dedicated effort to establish a benchmark 
dataset specifically designed for assessing the performance of KGQA systems 
over the MOF-KG. This effort leverages the exceptional 
capacity of ChatGPT to incorporate a diverse range of question types, 
domain-specific language, and complex relationships found within the MOF-KG.

%
%
\section{Developing a KGQA4MAT Benchmark Dataset over the MOF-KG}
\label{sec:benchmark}

We began developing our benchmark dataset by 
generating questions and corresponding Cypher queries
involving facts, comparison, superlative,
and aggregation. We created 161 such questions. 
For each question, we asked ChatGPT to rephrase it in 3 different ways.
Finally, we have 644 questions.
We split the data into train and test sets by the ratio 80:20. As a result, the train set
contains 515 questions and queries and the test set contains 129 questions and queries.
In the process of translating a question to a query, a KGQA system should prioritize the logical 
steps underlying the query and treat MOF identifications as interchangeable variables. Consequently, 
to ensure that MOF identifications do not play a crucial role in the translation, we substituted 
the MOF-related identifications in the test set with different ones from the training set.

\begin{figure}[!ht]
	\centering
	\includegraphics[width=1\textwidth]{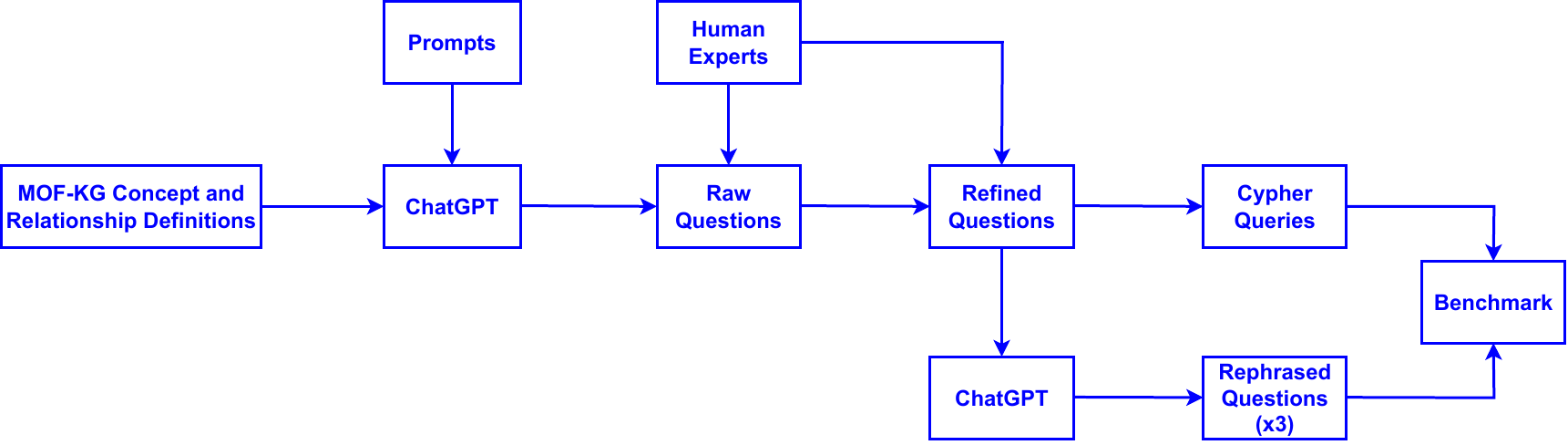}
	\caption{The process of generating the KGQA4MAT benchmark}
	\label{fig:benchmark-generation-process}
\end{figure}

Figure \ref{fig:benchmark-generation-process} illustrates the process of generating the benchmark.
First, the concepts and relationships within the MOF-KG ontology are incorporated into 
the prompt provided to ChatGPT. Example definitions are: \\
\texttt{
"MOF":["refcode":ID, "name", "refcode"]; \\ 
"Publication":["doi":ID, "year", "first\_page", "volume"]; \\
"Synthesis\_Procedure":["syn\_id":ID, "name", "symbol", "method"];\\
"(:MOF)-[:has\_publication]-(:Publication)";\\
"(:Synthesis\_Procedure)-[:has\_publication]-(:Publication)";\\
}
Next, ChatGPT is instructed to generate complex questions based on the MOF-KG definitions. 
Subsequently, human experts examine and revise the raw questions produced by ChatGPT to 
create a set of polished questions. In the fourth step, knowledge graph experts convert 
the refined questions into corresponding Cypher queries for the MOF-KG. Following this, 
ChatGPT is prompted once more to rephrase the refined questions in at least three distinct 
ways. Lastly, the original questions, their variations, and the corresponding Cypher 
queries collectively constitute the benchmark dataset.

%
%
\section{Leveraging ChatGPT to Translate User Questions to KG Queries}
\label{sec:approach}

ChatGPT, a powerful large language model (LLM), has achieved remarkable progress in the 
domain of natural language understanding. Owing to its extensive coverage of 
resources such as Wikipedia and its capacity to support natural language question 
answering, ChatGPT has attracted attention as a potential alternative to 
traditional knowledge-based question answering (KBQA) models. 
However, the utilization of ChatGPT for translating user 
questions into queries on specific knowledge graphs still 
requires extensive evaluation.

In order to assess ChatGPT's performance in understanding user questions and generating 
Cypher queries, we propose an approach that employs zero-shot, few-shot, and chain-of-thought \cite{chainOfThought} 
strategies. This approach aims to provide a comprehensive evaluation of ChatGPT's 
effectiveness in addressing the complexities associated with KGQA and facilitating its application 
to domain-specific knowledge graphs.

For few-shot learning, the approach needs to find the most semantically similar training questions for 
a test question. We leverage OpenAI's \texttt{text-embedding} \texttt{-ada-002} embedding model to embed
all the training and test questions. We then compute the cosine similarity between 
two embedding vectors for the similarity between the two associated questions. 
For chain-of-thought \cite{chainOfThought} reasoning, 
we first instruct ChatGPT to generate the chain of thought for a given Cypher query, then
refine the reasoning steps manually. Our prompt is as follows:
\begin{table}
\centering
\begin{tabular}{ll}
        \textbf{Promt:} & \texttt{Explain the reasoning in the Cypher query in logical steps.} \\
                        & \texttt{Use normal words in the explanation. No comments. Output} \\
                        & \texttt{the explanation only.}\\
\end{tabular}
\end{table}

\noindent
For example, given the Cypher query answering the question: \emph{In what order were the synthetic 
operations performed to create MOF-3?}, the chain-of-thought explanations are as follows:
\begin{table}
\centering
\begin{tabular}{ll}
        1. & \texttt{Find all MOFs that have a relation "has\_publication" to a "Publication".} \\
        2. & \texttt{Find all synthesis procedures that have a "has\_publication" relation to}\\ 
           & \texttt{the same "Publication".} \\
        3. & \texttt{Find all synthesis operations that are part of the previously found} \\ 
           & \texttt{synthesis procedures.} \\
        4. & \texttt{Filter the MOFs to only include those whose name contains "MOF-3".} \\
        5. & \texttt{Output the synthesis procedure id, name of the operation, and the order} \\
           & \texttt{of the operation.} \\
        6. & \texttt{Sort the output by the synthesis procedure id in ascending order,} \\ 
           & \texttt{followed by the operation order in ascending order.} \\
\end{tabular}
\end{table}

After pre-processing the train and test questions and queries, We proposed an approach 
of leveraging ChatGPT for KGQA with the following methods:
\begin{itemize}
    \item \textbf{Method 0: Answering questions directly.} Ask ChatGPT to answer the users question directly
    without user-provided additional information.
    \item \textbf{Method 1: Zero-shot learning from ontology only.} In this method, we prompt
ChatGPT with the definitions in an ontology. We then instruct ChatGPT to translate
a user question to a Cypher query based on the ontology.
    \item \textbf{Method 2: few-shot learning from a few pairs of train question and query.} Using the embeddings of 
    the test and train questions to find a few highly similar train questions to the test question. Prompt 
    ChatGPT with the pairs of matched train question and query. Instruct ChatGPT to translate the test question 
    to a Cypher query.
    \item \textbf{Method 3: few-shot learning from an ontology and a few pairs of train question and query.} As 
    in Method 2, include an ontology in the prompt, in addition to a few pairs of matched question and query. 
    \item \textbf{Method 4: few-shot learning from a few pairs of train question and query, and the chains-of-thought
    of the train queries.} As in Method 2, include the chains-of-thought of the train queries 
    in the prompt, in addition to the pairs of matched question and query.
    \item \textbf{Method 5: few-shot learning from an ontology and a few pairs of train question and query, 
    and the chains-of-thought of the train queries.} As in Method 3, include the chains-of-thought 
    of the train queries in the prompt, in addition to the definitions in an 
    ontology and the pairs of matched question and query.
\end{itemize}

\comment{
We prompt ChatGPT with either a specific domain ontology or
a few training examples. We then instruct ChatGPT to translate a user's question using the knowledge 
in the ontology or in the training examples. We do not directly ask ChatGPT to answer user's questions
treating ChatGPT itself as a comprehensive knowledge base as in \cite{evaluation-chatGPT-QA}. 
There are two main considerations. First, 
by instructing ChatGPT to use the knowledge in the given prompt, we hope to reduce the 'hallucination' 
issue that has besmirched ChatGPT. Second, the training data of ChatGPT may not cover the 
latest and specific terminology about the given domain. Instructing ChatGPT to use the prompted knowledge
would improve the accuracy of the KGQA translation. 
}

%
%
\section{Evaluation}
\label{sec:evaluation}

We evaluate the proposed approach on the KGQA4MAT benchmark. To assess
the generalizability of leveraging ChatGPT for KGQA, we
also evaluate the approach on the well-known QALD-9 dataset \cite{qald-9}.
We apply the OpenAI model \texttt{gpt-turbo-3.5} for ChatGPT batch processing. 
The underlying KG management system of the MOF-KG is
Neo4j, while the QALD-9 dataset is over a DBpedia SPARQL endpoint. Thus, 
our evaluation assesses the performance of
ChatGPT on translating user questions to both Cypher and SPARQL queries.

\subsection{Evaluation on the KGQA4MAT Benchmark}
\label{subsec:benchmark-evaluation}

Since MOF-KG is a domain specific knowledge graph, 
we evaluate the Method 1-5 in the proposed approach on KGQA4MAT benchmark
without asking ChatGPT to answer questions directly.
For each translated Cypher query, we execute it on the MOF-KG Neo4j database.
We then compare the results to that retrieved by the correct Cypher query
to see whether the translated query is correct.
We measure the performance using the precision, recall, and F1-score in terms of 
the correctness of the translated queries: 
\begin{equation}
    prcision = \frac{\text{\emph{\# of correctly translated queries}}}
{\text{\emph{\# of total translated queries}}}
\label{eq:precision}
\end{equation}

\begin{equation}
    recall = \frac{\text{\emph{\# of correctly translated queries}}}
{\text{\emph{\# of total correct queries}}}
\label{eq:recall}
\end{equation}

\begin{equation}
    F1\_score = 2(\frac{1}{precision} + \frac{1}{recall})^{-1}
    \label{eq:f1-score}
\end{equation}

For the few-shot methods, we tested using 1, 3, and 5 training examples. We did not see
significant differences between the results of using 1, 3, or 5 training examples. 
We used 1-shot learning to evaluate the 
full benchmark in order to conserve tokens when utilizing ChatGPT. 
Table \ref{tab:MOF-KG-question2query-results} shows the evaluation results.

\begin{table}[!ht]
    \centering
    \begin{tabular}{|l|c|}
    \hline
            &  \multicolumn{1}{|c|}{\textbf{}} \\
    \multicolumn{1}{|c|}{\textbf{Evaluation Method on the KGQA4MAT Benchmark}}  
    &   \multicolumn{1}{|c|}{\textbf{F1-Score}} \\
                    &  \multicolumn{1}{|c|}{\textbf{}} \\
    \hline

    \textbf{Method 1: Zero-shot learning from MOF-KG ontology only.} & 0.411 \\
            Prompt ChatGPT with the MOF-KG ontology definitions. & \\   
            Instruct  ChatGPT to translate a user question to a  & \\
            Cypher query based on the ontology. & \\
    \hline
    \textbf{Method 2: 1-shot learning from a pair of train question} & 0.829 \\
    \textbf{and query.} Using the embeddings of the test and train & \\
            questions to find the most similar train question to & \\
            the test question. Prompt ChatGPT with the pair of matched & \\ 
            train question and query. Instruct ChatGPT to translate a & \\
            test question to a Cypher query. & \\
    \hline
    \textbf{Method 3: 1-shot learning from the MOF-KG ontology and} & 0.845 \\ 
    \textbf{a pair of train question and query.} As in Method 2, & \\
            include MOF-KG ontology in the prompt, in addition to  & \\
            a pair of matched question and query. & \\
    \hline
    \textbf{Method 4: 1-shot learning from a pair of train question} & 0.876 \\ 
    \textbf{and query, and the chain-of-thought of the train query.} & \\
            As in Method 2, include the chain-of-thought of the train query & \\
            in the prompt, in addition to the pair of matched question and query & \\ 
    \hline
    \textbf{Method 5: 1-shot learning from the MOF-KG ontology and} & \textbf{0.891} \\
    \textbf{a pair of train question and query, and the chain-of-thought} & \\
    \textbf{of the train query.} As in Method 3, include the chain-of-thought & \\ 
            of the train query in the prompt, in addition to MOF-KG ontology & \\
            and the pair of matched question and query. & \\
     \hline
    \end{tabular}
    \caption{Evaluation results of leveraging ChatGPT to convert
    natural language questions to Cypher queries on the KGQA4MAT benchmark}
    \label{tab:MOF-KG-question2query-results}
\end{table}

Because we translated one query for each test question, the total number of translated questions equals 
to the total number of correct questions. As a result, $precision=reall=F1\_score$.
The results demonstrate that our approach effectively translates natural language 
questions into Cypher queries, achieving a
high F1-score of 0.891 when the ontology, a matched pair of $<$question, query$>$ 
as a training example, and a set of logical steps in a chain-of-thought explanation 
about the training query are provided.

\subsection{Evaluation on QALD-9 Dataset}
\label{subsec:qald-9-evaluation}

We also evaluate the performance of the proposed approach 
leveraging ChatGPT for KGQA on the 
QALD-9 \cite{qald-9} dataset, 
which contains natural language questions and corresponding SPARQL queries over DBpedia. 
QALD-9 is a multilingual question answering benchmark containing
408 train questions and 150 test questions. We only evaluate its English version of 
train and test questions. 
As a pre-trained large language model, ChatGPT has been trained on 
a variety of text sources, including Wikipedia. 
Consequently, when assessing ChatGPT on the QALD-9 dataset, we did 
not specifically incorporate the concept and relationship 
definitions from the DBpedia ontology in the prompts, assuming 
that ChatGPT has already learned the ontology.

Translated SPARQL queries are executed on the current DBpedia SPARQL 
endpoint\footnote{https://dbpedia.org/sparql}. Subsequently, we 
compare the results with those obtained from the correct SPARQL 
query to determine the accuracy of a translated query. The performance 
is measured using precision, recall, and F1-score, as previously defined 
in equations (\ref{eq:precision}), (\ref{eq:recall}), and (\ref{eq:f1-score}).

Table \ref{tab:QALD-9-question2query-results} shows the evaluation results
in comparison to the best performance in the KGQA leaderboard
on QALD-9\footnote{https://github.com/KGQA/leaderboard/blob/gh-pages/dbpedia/qald.md\#qald-9}
as of April, 2023.
The results show that the proposed approach that leverages
ChatGPT can successfully translate natural 
language questions into SPARQL queries with a competitive performance 
compared to other state-of-the-art approaches.

\begin{table}[!ht]
    \centering
    \begin{tabular}{|l|c|}
    \hline
            &  \multicolumn{1}{|c|}{\textbf{}} \\
    \multicolumn{1}{|c|}{\textbf{Evaluation Method on the QALD-9 Benchmark}}  
    &   \multicolumn{1}{|c|}{\textbf{F1-Score}} \\
                    &  \multicolumn{1}{|c|}{\textbf{}} \\
    \hline
    \textbf{Baseline 1: SGPT\_Q,K, 2022}, top 1 in the KGQA Leaderboard  & 0.67 \\
     as of April 2023 \cite{SGPT-generative} & \\
     \hline
     \textbf{Baseline 2: SGPT\_Q, 2022}, top 2 in the KGQA Leaderboard  & 0.60 \\
     as of April 2023 \cite{SGPT-generative} & \\
     \hline
     \textbf{Baseline 3: Stage I No Noise, 2022}, top 3 in the KGQA  & 0.55\\
     Leaderboard as of April 2023 \cite{KGQA-silhouette} & \\
                                                     
     \hline
     \textbf{Baseline 4: GPT-3.5v3, 2023}, top 6 in the KGQA Leaderboard  & 0.46 \\
     as of April 2023 \cite{evaluation-chatGPT-QA} & \\
     \hline
     \textbf{Baseline 5: ChatGPT, 2023}, top 7 in the KGQA Leaderboard  & 0.45 \\
     as of April 2023 \cite{evaluation-chatGPT-QA} & \\
    \hline
    \hline
    \textbf{Method 0: Answer the QALD-9 questions directly.} & 0.33 \\
            Instruct  ChatGPT to answer the QALD-9 
            questions directly & \\
    \hline
    \textbf{Method 1: Translate QALD-9 questions directly.} & 0.3 \\ 
            Instruct ChatGPT to translate QALD-9 questions & \\
            directly, and then query DBpedia using the translated & \\
            queries. & \\
    \hline
    \textbf{Method 2: 1-shot learning from a pair of train question} & 0.40 \\
    \textbf{and query.} Using the embeddings of the test and train & \\
            questions to find the most similar train question to & \\
            the test question. Prompt ChatGPT with the pair of matched & \\ 
            train question and query. Instruct ChatGPT to translate a & \\
            test question to a SPARQL query over DBpedia. & \\
    \hline
    \textbf{Method 3: 1-shot learning from a pair of train question} & \textbf{0.66} \\ 
    \textbf{and query, and the chain-of-thought of the train query.} & \\
            As in Method 2, include the chain-of-thought of the train query & \\
            in the prompt, in addition to the pair of matched question and query & \\ 
     \hline
    \end{tabular}
    \caption{Evaluation results of leveraging ChatGPT to convert
    natural language questions to SPARQL queries on the QALD-9 benchmark}
    \label{tab:QALD-9-question2query-results}
\end{table}

%
%
\section{Discussion}
\label{sec:discussion}

Our results demonstrate the potential of using ChatGPT to translate natural language questions 
to knowledge graph queries. As seen in the evaluation, the highest F1-score of 0.891 is achieved 
when the ontology, a matched pair of $<$question, query$>$ as a training example, and a set of 
logical steps in a chain-of-thought explanation about the query are provided.
Comparing the performance of ChatGPT on MOF-KG and QALD-9 datasets, we observe that our approach is 
generalizable and can be adapted to different knowledge domains. However, there is a significant 
gap between the best performance on MOF-KG (F1-score=0.891) and 
the best performance on QALD-9 (F1-score=0.66). One possible reason is the different sizes of 
the ontologies describing the two knowledge graphs. 
The MOF-KG ontology only has 26 concepts and relationships, while the DBpedia 
ontology currently covers 685 classes and 
2,795 different properties\footnote{http://wikidata.dbpedia.org/services-resources/ontology}. 
In our experiments, we relied on the ChatGPT's internal knowledge about the DBpedia 
ontology to translate users' questions. 
Despite utilizing only a single pair of $<$question, query$>$ as a training example with a 
chain-of-thought explanation, ChatGPT managed to achieve performance comparable to 
the leading systems on the current KGQA leaderboard. These top systems typically employ 
sophisticated data processing strategies and advanced deep-learning models.

We carefully analyzed the natural language (NL) questions for which ChatGPT made errors when translating 
them into knowledge graph (KG) queries.  
We observed that the primary mistakes occurred when ChatGPT failed to utilize the correct path 
within the knowledge graphs, including alternative paths.
For example, ChatGPT displayed a tendency to generate new concepts and properties in the KG queries, 
even though we explicitly incorporated the following instruction in the prompt: 
"\emph{only use the concepts and relationships in the given ontology.}" 
The another example is the `UNION' operation that combines multiple alternative paths. 
ChatGPT could not correctly apply `UNION' in many cases.
Here are some examples to further illustrate these issues.\\

\noindent
\textbf{Example from the KGQA4MAT Benchmark:} 

\noindent
\textbf{Question:} \texttt{What is the total number of operation conditions of all \\ synthesis operations for any MOFs `MIL-53'?}
\\

\noindent
\textbf{Training Question:} \texttt{What is the overall count of operation conditi- \\ ons for all the  
synthesis procedures associated with any MOFs  \\ `MOF-3'?}
\\

\noindent
\textbf{Training Query:} \texttt{MATCH (m:MOF)-[:has\_publication]->(:Publication) \\ <-
[:has\_publication]-(s:Synthesis\_Procedure)-[:has\_operation]-> \\ (o:Synthesis\_Operation)-
[:has\_condition]-> \\ (c:Synthesis\_Operation\_Condition) WHERE m.name CONTAINS 
`MOF-3' \\ RETURN count(c);}
\\

\noindent
\textbf{ChainOfThought:} \texttt{\{logical steps in the training query. Check 
\\ the detailed chain of thought steps in the training data set.\}}
\\

\noindent
\textbf{Correct Query:} \texttt{MATCH (m:MOF)-[:has\_publication]->(:Publication) \\ 
<-[:has\_publication]-(s:Synthesis\_Procedure)-[:has\_operation]-> \\ 
(o:Synthesis\_Operation)-[:has\_condition]-> \\ 
(c:Synthesis\_Operation\_Condition) WHERE m.name CONTAINS `MIL-53' \\ RETURN count(c)}
\\

\noindent
\textbf{\underline{Incorrect Answer:}} \texttt{MATCH (m:MOF)-[:has\_synthesisProcedure]-> \\ 
(sp:Synthesis\_Procedure)-[:has\_operation]->(o:Synthesis\_Operation) \\ 
-[:has\_condition]->(c:Synthesis\_Operation\_Condition) WHERE m.name \\ CONTAINS `MIL-53' RETURN count(c);}
\\

\noindent
\textbf{Issues:} Despite the training question being highly similar to the question being asked, 
with the correct query differing only in the MOF identification, ChatGPT still made an error in 
the translation process. It is important to note that ChatGPT was provided with chain-of-thought 
logical steps in the training query to facilitate its understanding. However, the model still 
introduced an invented relationship \texttt{(m:MOF)-[:has\_synthesisProcedure]->(sp:Synthesis\_Procedure)}, 
which ultimately led to an incorrect answer. 
\\

\noindent
\textbf{Example from the QALD-9 Benchmark:} 

\noindent
\textbf{Question:} \texttt{Which countries in the European Union adopted the Euro?}
\\

\noindent
\textbf{Training Question:} \texttt{Which countries adopted the Euro?}
\\

\noindent
\textbf{Training Query:} \\
\texttt{SELECT DISTINCT ?uri \\
WHERE \{ \\
?uri a <http://dbpedia.org/ontology/Country> \\
\{ ?uri <http://dbpedia.org/ontology/currency> \\ <http://dbpedia.org/resource/Euro> \} \\
UNION \\
\{ ?uri <http://dbpedia.org/property/currencyCode> "EUR"@en \} \\
}
\\

\noindent
\textbf{ChainOfThought:} \texttt{\{logical steps in the training query. Check \\ 
the detailed chain of thought steps in the training data set.\}}
\\

\noindent
\textbf{Correct Query:} \\
\texttt{SELECT DISTINCT ?uri \\ WHERE \{ ?uri a <http://dbpedia.org/ontology/Country> \\ \{ ?uri <http://dbpedia.org/ontology/currency> \\ <http://dbpedia.org/resource/Euro> \} \\ UNION \\ \{ ?uri <http://dbpedia.org/property/currencyCode> "EUR"@en \} \}}
\\

\noindent
\textbf{\underline{Incorrect Answer:}} \\
\texttt{PREFIX rdf: <http://www.w3.org/1999/02/22-rdf-syntax-ns\#>        \\     PREFIX rdfs: <http://www.w3.org/2000/01/rdf-schema\#>  \\           PREFIX dbo: <http://dbpedia.org/ontology/>        \\     PREFIX dct: <http://purl.org/dc/terms/>        \\     PREFIX dbp: <http://dbpedia.org/property/>          \\   SELECT DISTINCT ?country            \\ WHERE \{ ?country a dbo:Country ;  \\                       dct:subject <http://dbpedia.org/resource/Category: \\ Member\_states\_of\_the\_European\_Union> .  \\    ?country dbo:currency <http://dbpedia.org/resource/Euro> .             \}}
\\

\noindent
\textbf{Issues:} The training question closely resembles the test question, with the training query involving 
a `UNION' operation that combines two alternative paths. Despite the chain-of-thought steps providing 
instructions for the `UNION' operation, ChatGPT failed to generate the correct query, 
resulting in a missing 'UNION' operation.

These examples demonstrate the challenges that needs to be addressed in order to 
improve the approach of utilizing ChatGPT in generating accurate KG queries from NL questions.

%
%
\section{Conclusion}
\label{sec:conclusion}

We developed a benchmark dataset 
for Knowledge Graph Question Answering for Materials Science (KGQA4MAT) 
and a systematic approach for utilizing ChatGPT to translate natural language 
questions into formal KG queries. Our evaluation demonstrated the potential of
utilizing ChatGPT for  
addressing KGQA issues across different platforms and query languages.
The MOF-KG, KGQA4MAT benchmark, and the proposed approach facilitate
further development on user-friendly interfaces enabling 
domain experts to more easily access and interrogate the wealth of information contained 
within materials science knowledge graphs. Future research directions include 
refining the proposed approach, 
expanding the benchmark dataset, and exploring alternative techniques for KGQA in 
the context of materials science.

\subsubsection{Benchmark Repository.}
The benchmark data, MOF-KG definitions, ChatGPT prompts, Python code for 
evaluation are accessible via the following GitHub repository:
\texttt{https://github.com/kgqa4mat/KGQA4MAT}\footnote{https://github.com/kgqa4mat/KGQA4MAT} .

\subsubsection{Acknowledgments.}
This project is partially supported by the Drexel Office of Faculty Affairs’ 2022 Faculty 
Summer Research awards \#284213, and the U.S. National Science Foundation Office of 
Advanced Cyberinfrastructure (OAC) Grant \#1940239,  \#1940307, and \#2118201.

\newpage
%
%
\bibliographystyle{unsrt}  

\begin{thebibliography}{10}
	
	\bibitem{reticular-chemistry-all-dimensions}
	Omar Yaghi.
	\newblock Reticular chemistry in all dimensions.
	\newblock {\em ACS Central Science}, 5(8):1295--1300, 8 2019.
	
	\bibitem{MOF-3}
	Mohamed Eddaoudi, Hailian Li, and O.~M. Yaghi.
	\newblock {Highly Porous and Stable Metal-Organic Frameworks: Structure Design
		and Sorption Properties}.
	\newblock {\em Journal of the American Chemical Society}, 122(7):1391--1397,
	2000.
	
	\bibitem{development-CSD-MOF}
	Peyman Moghadam, Aurelia Li, Seth Wiggin, Andi Tao, Andrew Maloney, Peter Wood,
	Suzanna Ward, and David Fairen-Jimenez.
	\newblock Development of a cambridge structural database subset: A collection
	of metal–organic frameworks for past, present, and future.
	\newblock {\em Chemistry of Materials}, 29(7):2618--2625, 3 2017.
	
	\bibitem{increasing-topoloical-diversity}
	Ryther Anderson and Diego Gómez-Gualdrón.
	\newblock Increasing topological diversity during computational “synthesis”
	of porous crystals: how and why.
	\newblock {\em CrystEngComm}, 21(10):1653--1665, 2019.
	
	\bibitem{understanding-diversity-MOF}
	Seyed~Mohamad Moosavi, Aditya Nandy, Kevin~Maik Jablonka, Daniele Ongari,
	Jon~Paul Janet, Peter~G. Boyd, Yongjin Lee, Berend Smit, and Heather~J.
	Kulik.
	\newblock Understanding the diversity of the metal-organic framework ecosystem.
	\newblock {\em Nature Communications}, 11(1):4068, 2020.
	
	\bibitem{DigiMOF}
	Kristian Gubsch, Rosalee Bence, Lawson Glasby, and Peyman Moghadam.
	\newblock {DigiMOF: A Database of MOF Synthesis Information Generated via Text
		Mining}.
	\newblock {\em Chemarxiv}, 4 2022.
	
	\bibitem{MOF-synthesis-automatic-dataMining}
	Yi~Luo, Saientan Bag, Orysia Zaremba, Adrian Cierpka, Jacopo Andreo, Stefan
	Wuttke, Pascal Friederich, and Manuel Tsotsalas.
	\newblock Mof synthesis prediction enabled by automatic data mining and machine
	learning.
	\newblock {\em Angewandte Chemie International Edition}, 61(19):e202200242,
	2022.
	
	\bibitem{mining-insights-MOF}
	Hyunsoo Park, Yeonghun Kang, Wonyoung Choe, and Jihan Kim.
	\newblock Mining insights on metal–organic framework synthesis from
	scientific literature texts.
	\newblock {\em Journal of Chemical Information and Modeling}, 62(5):1190--1198,
	2022.
	\newblock PMID: 35195419.
	
	\bibitem{kg-empowered-materials}
	Xintong Zhao, Jane Greenberg, Scott McClellan, Yong-Jie Hu, Steven Lopez,
	Semion~K Saikin, Xiaohua Hu, and Yuan An.
	\newblock Knowledge graph-empowered materials discovery.
	\newblock In {\em 1st Workshop on Knowledge Graph and Big Data collocated with
		2021 IEEE International Conference on Big Data (Big Data)}, 2021.
	
	\bibitem{mof-kg}
	Yuan An, Jane Greenberg, Xintong Zhao, Xiaohua Hu, Scott McCLellan, Alex
	Kalinowski, Fernando~J. Uribe-Romo, Kyle Langlois, Jacob Furst, Diego~A.
	Gómez-Gualdrón, Fernando Fajardo-Rojas, and Katherine Ardila.
	\newblock Building open knowledge graph for metal-organic frameworks (mof-kg):
	Challenges and case studies.
	\newblock In {\em International Workshop on Knowledge Graphs \& Open Knowledge
		Network (OKN) Co-located with the 28th ACM SIGKDD Conference}, Washington,
	DC, August 15, 2022, Washington, DC.
	
	\bibitem{mof-lama}
	Yuan An, Jane Greenberg, Xiaohua Hu, Alex Kalinowski, Xiao Fang, Xintong Zhao,
	Scott McClellan, Fernando~J. Uribe-Romo, Diego~A. Gómez-Gualdrón, Kyle
	Langlois, Jacob Furst, Fernando Fajardo-Rojas, Katherine Ardila, Semion~K.
	Saikin, Corey A.~Harper Harper, and Ron Daniel.
	\newblock Exploring pre-trained language models to build knowledge graph for
	metal-organic frameworks (mofs).
	\newblock In {\em In 2nd Workshop on Knowledge Graph and Big Data collocated
		with IEEE BigData Conference}, 2022.
	
	\bibitem{qald-9}
	Ricardo Usbeck, Ria~Hari Gusmita, Axel-Cyrille~Ngonga Ngomo, and Muhammad
	Saleem.
	\newblock 9th challenge on question answering over linked data (qald-9)
	(invited paper).
	\newblock In {\em Semdeep/NLIWoD@ISWC}, 2018.
	
	\bibitem{MMKG}
	Xiaoming Zhang, Xin Liu, Xin Li, and Dongyu Pan.
	\newblock Mmkg: An approach to generate metallic materials knowledge graph
	based on dbpedia and wikipedia.
	\newblock {\em Computer Physics Communications}, 211:98--112, 2017.
	\newblock High Performance Computing for Advanced Modeling and Simulation of
	Materials.
	
	\bibitem{propnet}
	David Mrdjenovich, Matthew Horton, Joseph Montoya, Christian Legaspi, Shyam
	Dwaraknath, Vahe Tshitoyan, Anubhav Jain, and Kristin Persson.
	\newblock propnet: A knowledge graph for materials science.
	\newblock {\em Matter}, 2(2):464--480, 2 2020.
	
	\bibitem{Nanomine}
	Jamie Mccusker, Neha Keshan, Sabbir Rashid, Michael Deagen, Cate Brinson, and
	Deborah Mcguinness.
	\newblock {NanoMine: A Knowledge Graph for Nanocomposite Materials Science}.
	\newblock {\em Lecture Notes in Computer Science}, pages 144--159, 2020.
	
	\bibitem{MatKG}
	Vineeth Venugopal, Sumit Pai, and Elsa Olivetti.
	\newblock Matkg: The largest knowledge graph in materials science -- entities,
	relations, and link prediction through graph representation learning.
	\newblock In {\em In AI4Mat workshop in NeurIPS 2022}, 2022.
	
	\bibitem{survey-QA-SW}
	Marta Sabou, Konrad H\"{o}ffner, Sebastian Walter, Edgard Marx, Ricardo Usbeck,
	Jens Lehmann, and Axel-Cyrille Ngonga~Ngomo.
	\newblock Survey on challenges of question answering in the semantic web.
	\newblock {\em Semant. Web}, 8(6):895–920, jan 2017.
	
	\bibitem{LC-QuAD}
	Priyansh Trivedi, Gaurav Maheshwari, Mohnish Dubey, and Jens Lehmann.
	\newblock Lc-quad: A corpus for complex question answering over knowledge
	graphs.
	\newblock In {\em International Semantic Web Conference}, 2017.
	
	\bibitem{LC-QuAD-2}
	Mohnish Dubey, Debayan Banerjee, Abdelrahman Abdelkawi, and Jens Lehmann.
	\newblock Lc-quad 2.0: A large dataset for complex question answering over
	wikidata and dbpedia.
	\newblock In {\em International Semantic Web Conference}, page 69–78.
	Springer-Verlag, 2019.
	
	\bibitem{SciQA}
	Pan Lu, Swaroop Mishra, Tony Xia, Liang Qiu, Kai-Wei Chang, Song-Chun Zhu,
	Oyvind Tafjord, Peter Clark, and Ashwin Kalyan.
	\newblock Learn to explain: Multimodal reasoning via thought chains for science
	question answering.
	\newblock In {\em The 36th Conference on Neural Information Processing Systems
		(NeurIPS)}, 2022.
	
	\bibitem{what-is-KGQA-benchmark}
	N.~Steinmetz and KU. Sattler.
	\newblock {What is in the KGQA Benchmark Datasets? Survey on Challenges in
		Datasets for Question Answering on Knowledge Graphs}.
	\newblock {\em J Data Semant}, 10:241–265, 2021.
	
	\bibitem{comparative-survey-NL-databases}
	Katrin Affolter, Kurt Stockinger, and Abraham Bernstein.
	\newblock A comparative survey of recent natural language interfaces for
	databases.
	\newblock {\em The VLDB Journal}, 28:793 -- 819, 2019.
	
	\bibitem{zou14natural}
	Lei Zou, Ruizhe Huang, Haixun Wang, Jeffrey~Xu Yu, Wenqiang He, and Dongyan
	Zhao.
	\newblock Natural language question answering over rdf: a graph data driven
	approach.
	\newblock In {\em Proceedings of the 2014 ACM SIGMOD international conference
		on Management of data}, pages 313--324, 2014.
	
	\bibitem{deep-neural-approach-KGQA}
	Sukannya Purkayastha, Saswati Dana, Dinesh Garg, Dinesh Khandelwal, and
	G.P~Shrivatsa Bhargav.
	\newblock A deep neural approach to kgqa via sparql silhouette generation.
	\newblock In {\em 2022 International Joint Conference on Neural Networks
		(IJCNN)}, pages 1--8, 2022.
	
	\bibitem{learning-to-rank-QG}
	Gaurav Maheshwari, Priyansh Trivedi, Denis Lukovnikov, Nilesh Chakraborty, Asja
	Fischer, and Jens Lehmann.
	\newblock Learning to rank query graphs for complex question answering over
	knowledge graphs.
	\newblock In Chiara Ghidini, Olaf Hartig, Maria Maleshkova, Vojt{\v{e}}ch
	Sv{\'a}tek, Isabel Cruz, Aidan Hogan, Jie Song, Maxime Lefran{\c{c}}ois, and
	Fabien Gandon, editors, {\em The Semantic Web -- ISWC 2019}, pages 487--504,
	Cham, 2019. Springer International Publishing.
	
	\bibitem{querying-KG-NL}
	Shiqi Liang, Kurt Stockinger, Tarcisio~Mendes de~Farias, Maria Anisimova, and
	Manuel Gil.
	\newblock Querying knowledge graphs in natural language.
	\newblock {\em Journal of Big Data}, 8(1):3, 2021.
	
	\bibitem{SGPT-generative}
	Md~Rashad Al~Hasan Rony, Uttam Kumar, Roman Teucher, Liubov Kovriguina, and
	Jens Lehmann.
	\newblock Sgpt: A generative approach for sparql query generation from natural
	language questions.
	\newblock {\em IEEE Access}, 10:70712--70723, 2022.
	
	\bibitem{Fang-prompt-design}
	Xiao Fang, Alexander Kalinowski, Haoran Zhao, Ziao You, Yuhao Zhang, and Yuan
	An.
	\newblock Prompt design and answer processing for knowledge base construction
	from pre-trained language models (lm-kbc).
	\newblock In {\em the Knowledge Base Construction from Pre-trained Language
		Models (LM-KBC) Challenge @ 21st International Semantic Web Conference},
	2022.
	
	\bibitem{can-chatgpt-understand}
	Qihuang {Zhong}, Liang {Ding}, Juhua {Liu}, Bo~{Du}, and Dacheng {Tao}.
	\newblock {Can ChatGPT Understand Too? A Comparative Study on ChatGPT and
		Fine-tuned BERT}.
	\newblock {\em arXiv e-prints}, page arXiv:2302.10198, February 2023.
	
	\bibitem{how-robust-gpt-3.5}
	Xuanting {Chen}, Junjie {Ye}, Can {Zu}, Nuo {Xu}, Rui {Zheng}, Minlong {Peng},
	Jie {Zhou}, Tao {Gui}, Qi~{Zhang}, and Xuanjing {Huang}.
	\newblock {How Robust is GPT-3.5 to Predecessors? A Comprehensive Study on
		Language Understanding Tasks}.
	\newblock {\em arXiv e-prints}, page arXiv:2303.00293, March 2023.
	
	\bibitem{chatgpt-versus-traditional}
	Reham {Omar}, Omij {Mangukiya}, Panos {Kalnis}, and Essam {Mansour}.
	\newblock {ChatGPT versus Traditional Question Answering for Knowledge Graphs:
		Current Status and Future Directions Towards Knowledge Graph Chatbots}.
	\newblock {\em arXiv e-prints}, page arXiv:2302.06466, February 2023.
	
	\bibitem{evaluation-chatGPT-QA}
	Yiming {Tan}, Dehai {Min}, Yu~{Li}, Wenbo {Li}, Nan {Hu}, Yongrui {Chen}, and
	Guilin {Qi}.
	\newblock {Evaluation of ChatGPT as a Question Answering System for Answering
		Complex Questions}.
	\newblock {\em arXiv e-prints}, page arXiv:2303.07992, March 2023.
	
	\bibitem{RCSR}
	Michael O’keeffe, Maxim Peskov, Stuart Ramsden, and Omar Yaghi.
	\newblock {The Reticular Chemistry Structure Resource (RCSR) Database of, and
		Symbols for, Crystal Nets}.
	\newblock {\em Accounts of Chemical Research}, 41(12):1782--1789, 12 2008.
	
	\bibitem{terminology-of-MOF}
	Stuart Batten, Neil Champness, Xiao-Ming Chen, Javier Garcia-Martinez, Susumu
	Kitagawa, Lars Öhrström, Michael O’keeffe, Myunghyun Paik~Suh, and Jan
	Reedijk.
	\newblock {Terminology of metal–organic frameworks and coordination polymers
		(IUPAC Recommendations 2013)}.
	\newblock {\em Pure and Applied Chemistry}, 85(8):1715--1724, 7 2013.
	
	\bibitem{Ashino-materials-ontology}
	Toshihiro Ashino.
	\newblock Materials ontology: An infrastructure for exchanging materials
	information and knowledge.
	\newblock {\em Data Science Journal}, 9:54--61, 2010.
	
	\bibitem{ChEBI}
	K~Degtyarenko, P~De~Matos, M~Ennis, J~Hastings, M~Zbinden, A~Mcnaught,
	R~Alcantara, M~Darsow, M~Guedj, and M~Ashburner.
	\newblock Chebi: a database and ontology for chemical entities of biological
	interest.
	\newblock {\em Nucleic Acids Research}, 36(Database):D344--D350, 12 2007.
	
	\bibitem{EMMO}
	{EMMO}.
	\newblock {Elementary Multiperspective Material Ontology (EMMO)}.
	\newblock {\em https://github.com/emmo-repo/EMMO}, 2022.
	
	\bibitem{MDO}
	Huanyu Li, Rickard Armiento, and Patrick Lambrix.
	\newblock {\em An Ontology for the Materials Design Domain}, pages 212--227.
	\newblock Lecture Notes in Computer Science. Springer International Publishing,
	2020.
	
	\bibitem{NIST-MaterialsOntology}
	Andrea Medina-Smith, Chandler Becker, Raymond Plante, Laura Bartolo, Alden
	Dima, James Warren, and Robert Hanisch.
	\newblock A controlled vocabulary and metadata schema for materials science
	data discovery.
	\newblock {\em Data Science Journal}, 20, 2021.
	
	\bibitem{MatPortal}
	{MatPortal}.
	\newblock {MatPortal}.
	\newblock {\em https://matportal.org/}.
	
	\bibitem{CoREMOF}
	Yongchul Chung, Emmanuel Haldoupis, Benjamin Bucior, Maciej Haranczyk, Seulchan
	Lee, Hongda Zhang, Konstantinos Vogiatzis, Marija Milisavljevic, Sanliang
	Ling, Jeffrey Camp, Ben Slater, J~Siepmann, David Sholl, and Randall Snurr.
	\newblock Advances, updates, and analytics for the computation-ready,
	experimental metal–organic framework database: Core mof 2019.
	\newblock {\em Journal of Chemical and Engineering Data}, 64(12):5985--5998, 11
	2019.
	
	\bibitem{targeted-classification-MOFs}
	Peyman~Z. Moghadam, Aurelia Li, Xiao-Wei Liu, Rocio Bueno-Perez, Shu-Dong Wang,
	Seth~B. Wiggin, Peter~A. Wood, and David Fairen-Jimenez.
	\newblock Targeted classification of metal–organic frameworks in the
	cambridge structural database (csd).
	\newblock {\em Chem. Sci.}, 11:8373--8387, 2020.
	
	\bibitem{identification-schemes-MOFs}
	Benjamin~J. Bucior, Andrew~S. Rosen, Maciej Haranczyk, Zhenpeng Yao, Michael~E.
	Ziebel, Omar~K. Farha, Joseph~T. Hupp, J.~Ilja Siepmann, Alán Aspuru-Guzik,
	and Randall~Q. Snurr.
	\newblock Identification schemes for metal–organic frameworks to enable rapid
	search and cheminformatics analysis.
	\newblock {\em Crystal Growth \& Design}, 19(11):6682--6697, 2019.
	
	\bibitem{chainOfThought}
	Jason Wei, Xuezhi Wang, Dale Schuurmans, Maarten Bosma, brian ichter, Fei Xia,
	Ed~H. Chi, Quoc~V Le, and Denny Zhou.
	\newblock Chain of thought prompting elicits reasoning in large language
	models.
	\newblock In Alice~H. Oh, Alekh Agarwal, Danielle Belgrave, and Kyunghyun Cho,
	editors, {\em Advances in Neural Information Processing Systems}, 2022.
	
	\bibitem{KGQA-silhouette}
	Sukannya Purkayastha, Saswati Dana, Dinesh Garg, Dinesh Khandelwal, and
	G.P~Shrivatsa Bhargav.
	\newblock A deep neural approach to kgqa via sparql silhouette generation.
	\newblock In {\em 2022 International Joint Conference on Neural Networks
		(IJCNN)}, pages 1--8, 2022.
	
\end{thebibliography}

\end{document}